\documentclass[runningheads]{llncs}
\usepackage{graphicx}
\usepackage{amsmath,amssymb} 
\usepackage{color}
\usepackage{caption} 
\usepackage{booktabs}

\begin{document}

\title{Augmentation for small object detection} 
\titlerunning{Augmentation for small object detection}


\author{Mate Kisantal \inst{1} \email{kisantal.mate@gmail.com} \\
Zbigniew Wojna \inst{1,2} \email{zbigniewwojna@gmail.com} \\
Jakub Murawski \inst{2,3} \email{kuba.murawski96@gmail.com} \\
Jacek Naruniec \inst{3} \email{j.naruniec@ire.pw.edu.pl} \\
Kyunghyun Cho \inst{4} \email{kyunghyun.cho@nyu.edu}}
%

\authorrunning{M. Kisantal et al.} 


\institute{Tensorflight, Inc. \and
University College London \and
Warsaw University of Technology \and
New York University}

\maketitle

\begin{abstract}
In the recent years, object detection has experienced impressive progress. Despite these improvements, there is still a significant gap in the performance between the detection of small and large objects. We analyze the current state-of-the-art model, Mask-RCNN, on a challenging dataset, MS COCO. We show that the overlap between small ground-truth objects and the predicted anchors is much lower than the expected IoU threshold. We conjecture this is due to two factors; (1) only a few images are containing small objects, and (2) small objects do not appear enough even within each image containing them. 
We thus propose to oversample those images with small objects and augment each of those images by copy-pasting small objects many times. 
It allows us to trade off the quality of the detector on large objects with that on small objects. We evaluate different pasting augmentation strategies, and ultimately, we achieve 9.7\% relative improvement on the instance segmentation and 7.1\% on the object detection of small objects, compared to the current state of the art method on MS COCO. 

\end{abstract}

\section{Introduction}
\begin{figure}[t]
\centering
\includegraphics[width=0.7\textwidth]{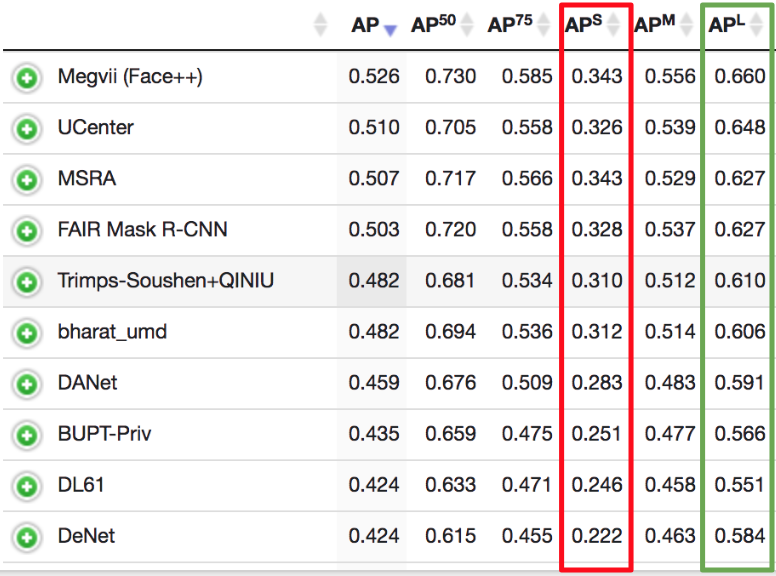}
\caption{In the top submissions for MS COCO instance segmentation challenge, the AP detection metric for small objects is 2-3 times lower than that for large objects.}
\label{fig:leaderboard}
\end{figure}

\begin{figure}[t]
\centering
\includegraphics[width=0.49\textwidth]{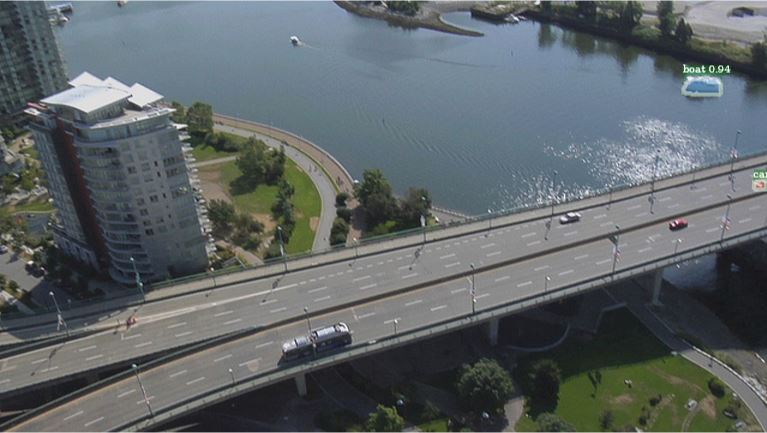}
\includegraphics[width=0.49\textwidth]{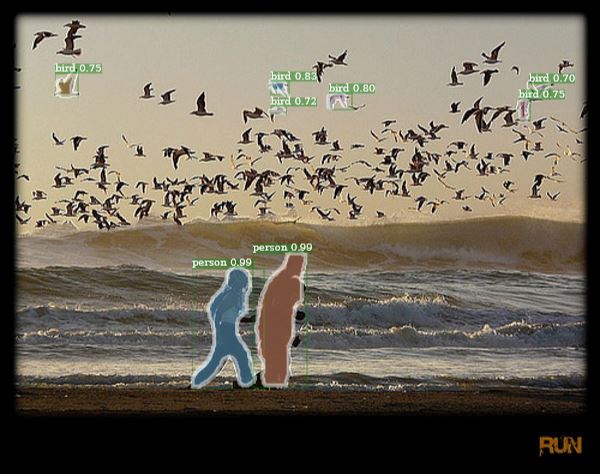}
\caption{Sample predictions from the Mask-RCNN, where many small objects were missed by the system, despite their clear visibility within a reasonable context. 
For instance, only six birds were detected out of hundreds of them.}
\label{fig:smallObject}
\end{figure}

Detecting objects in an image is one of the fundamental tasks of today’s computer vision research, as it is often
a starting point for many real world applications, including
robotics and 
self-driving cars, 
satellite and aerial image analysis, and
the localization of organs and masses in medical images. 
This important problem of object detection has recently experienced a lot of progress. The top-1 solution on MS COCO object detection competition,\footnote{
\url{http://cocodataset.org/\#detection-leaderboard}
} has progressed from the average precision (AP) of 0.373 in 2015 \cite{ren2015faster} to 0.525 
in 2017 (at IoU=.50:.05:.95 which is a primary challenge metric.) Similar progress can be observed in the instance segmentation problem in the context of MS COCO instance segmentation challenge. 
Despite these improvements, existing solutions often underperform with small objects, where small objects are defined as in Table~\ref{table:sizes} in the case of MS COCO.
It is evident from the significant gap in the performance between the detection of small and large objects. See for instance
Figure~\ref{fig:leaderboard} which lists the top ranking submissions for the MS COCO instance segmentation challenge. 
A similar issue is observed in the instance segmentation task as well. For instance, see the sample predictions from the current state-of-the-art model, Mask-RCNN, in Figure~\ref{fig:smallObject}, where the model has missed most of the small objects.

\begin{table}[t]
\centering
\begin{tabular}{ c | c  c }
              & Min rectangle area & Max rectangle area\\ 
              \toprule
Small object  & 0$\times$0        & 32$\times$32 \\ 
Medium object & 32$\times$32 & 96$\times$96 \\
Large object  & 96$\times$96 &  $\infty \times \infty$       \\ 
\end{tabular}
\caption{The definitions of the small, medium and large objects in MS COCO.}
\label{table:sizes}
\end{table}

Small object detection is crucial in many downstream tasks. 
Detecting small or distant objects in the high-resolution scene photographs from the car is necessary to deploy self-driving cars safely. Many objects, such as traffic signs \cite{deshmukh2013real,sermanet2011traffic} or pedestrians \cite{ouyang2013joint}, 
are often barely visible on the high-resolution images. In medical imaging, early detection of masses and tumors 
is crucial for making an accurate, early diagnosis, when such elements can easily be only a few pixels in size \cite{bottema2000detection,nagarajan2013classification}. Automatic industrial inspection can also benefit from small object detection by the localization of small defects that can be visible on the material surfaces \cite{abouelela2005automated,ng2006automatic}. Another application is satellite image analysis, where objects, such as cars, ships, and houses, must be effectively annotated  \cite{modegi2008small,kampffmeyer2016semantic}. With an average of 0.5-5m per pixel resolution, these objects are again just a few pixels in size.
In other words, 
small object detection and segmentation 
requires more attention, as more complex systems are being deployed in the real world. 
We, therefore, propose a new method to improve small object detection.

We focus on the state-of-the-art object detector, Mask R-CNN~\cite{he2017mask}, on a challenging dataset, MS COCO. We note two properties of this dataset regarding small objects. First, we observe that there are relatively fewer images that contain small objects in the dataset, which potentially biases any detection model to focus more on medium and large objects. Second, the area covered by small objects is much smaller, implying the lack of diversity in the locations of small objects. We conjecture this makes it difficult for the object detection model to generalize to small objects in the test time when they appear in less explored portions of an image.

\begin{figure}[t]
\centering
\includegraphics[width=\textwidth]{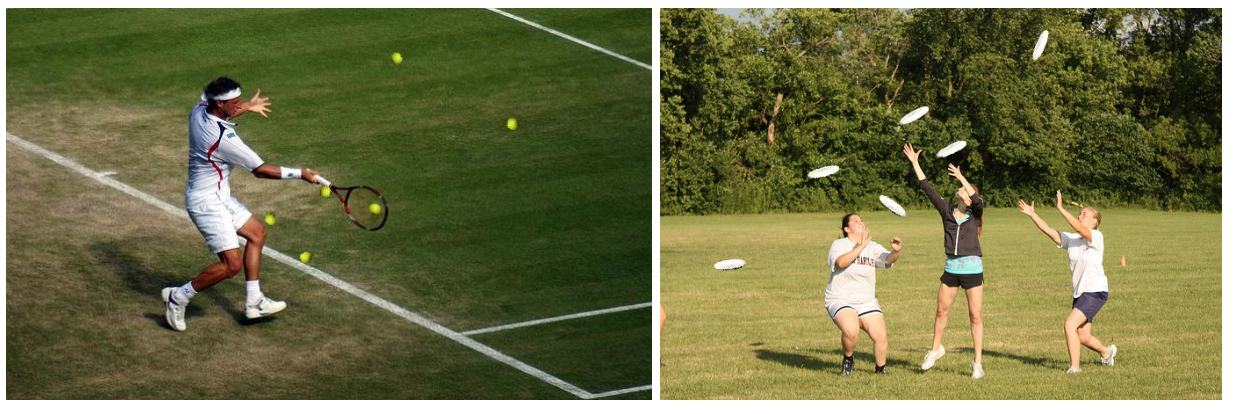}
\caption{Examples of artificial augmentation by copy pasting the small objects. As we can observe in these examples, pasting at the same image gives a high chance of the right surrounding context of the small object.}
\label{fig:augmentation}
\end{figure}

We tackle the first issue by oversampling those images containing small objects. The second issue is addressed by copy-pasting small objects multiple times in each image containing small objects. When pasting each object, we ensure that pasted objects do not overlap with any existing object. This increases the diversity in the locations of small objects while ensuring that those objects appear in correct context, as shown in Fig.~\ref{fig:augmentation}. The increase in the number of small objects in each image further addresses the 
issue of a small number of positively matched anchors, 
which we quantitatively analyze in Section~\ref{sec:analysis}. 
Overall, we achieve 9.7\% relative improvement for the instance segmentation and 7.1\% for object detection for small objects, compared to the current state-of-the-art method, Mask R-CNN, on MS COCO.

\begin{figure}[ht!]
\centering
\includegraphics[width=0.49\textwidth]{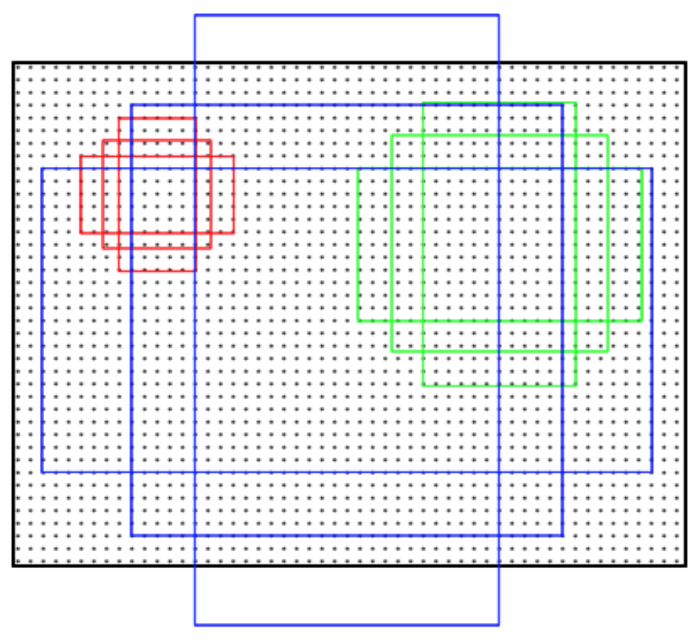}
\includegraphics[width=0.49\textwidth]{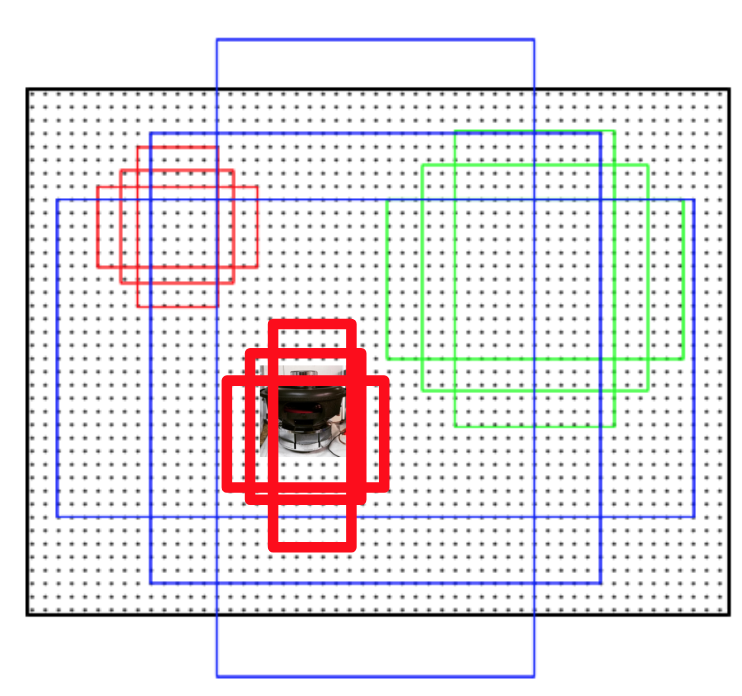}
\includegraphics[width=0.49\textwidth]{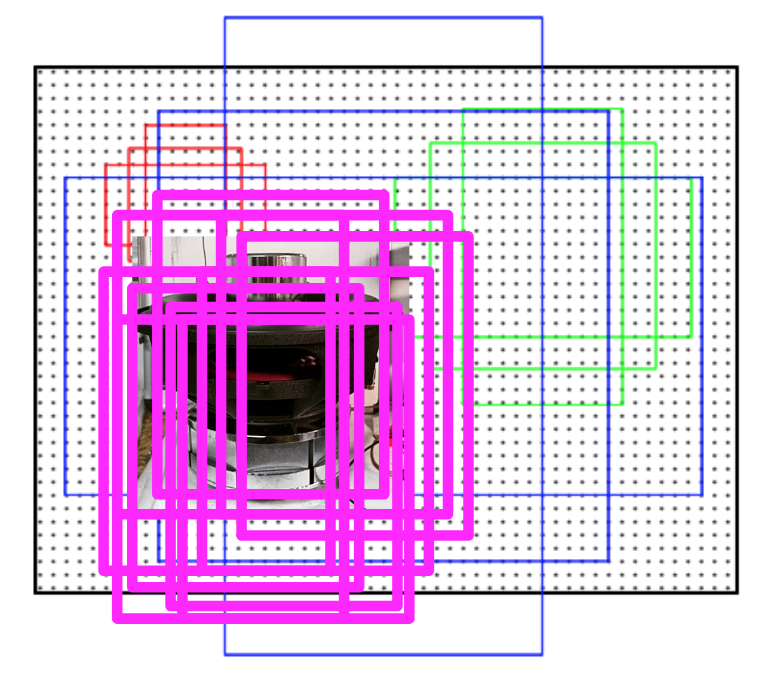}
\includegraphics[width=0.49\textwidth]{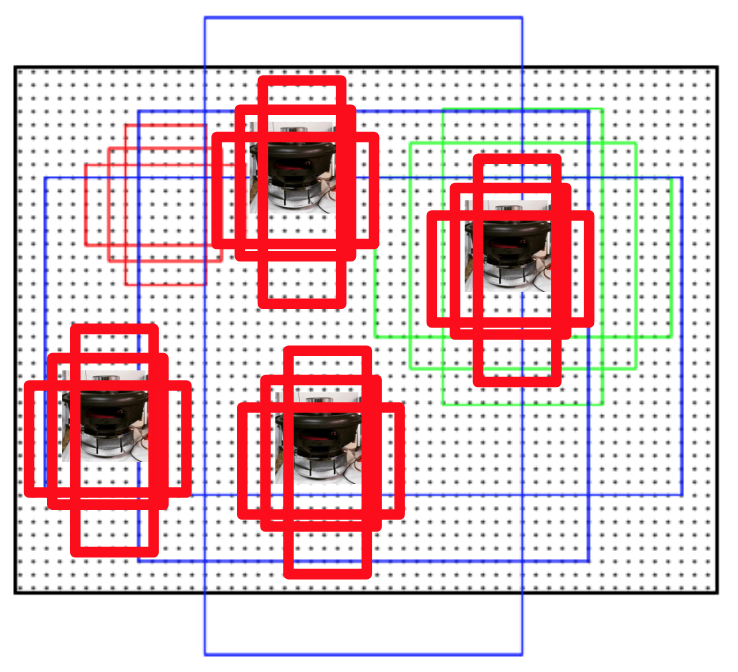}
\caption{Schematic illustration of anchors of different scales matching the ground truth objects. Small objects have much less anchors matched. To overcome the problem, we propose to artificially augment the images by copy-pasting small objects, so that there is more anchors positively matched with small objects during training. }
\label{fig:schematic_anchors}
\end{figure}

\section{Related Work}

\paragraph{Object Detection} 
Faster region-based convolutional neural network (Faster R-CNN) \cite{ren2015faster}, Region-based fully convolutional network (R-FCN) \cite{dai2016r} and Single Shot Detector (SSD) \cite{liu2016ssd} are three 
major approaches to object detection and they differ by whether and where the region proposal is attached \cite{huang2017speed}. Faster R-CNN and its variants 
are designed to help with a variety of object scales, as differential cropping merges all proposals into a single resolution. This, however, happens inside a deep convolutional network, and the resulting cropped boxes may not align perfectly with objects, which may hurt its performance in practice.
SSD was recently extended into Deconvolutional Single Shot Detector (DSSD) \cite{fu2017dssd}, that upsamples the low-resolution features of SSD by the transposed convolutions in the decoder part \cite{wojna2017devil}, to increase the internal spatial resolution. 
Similarly, Feature Pyramid Network (FPN) \cite{lin2017feature} extends the Faster R-CNN with decoder type sub-network. 

\paragraph{Instance Segmentation} 
Instance segmentation goes beyond object detection and requires 
predicting the exact mask of each object. 
Multi-Task Network Cascades (MNC) \cite{dai2016instance} build a cascade of prediction and mask refinement.
Fully convolutional instance-aware semantic segmentation (FCIS) \cite{li2016fully} is a fully convolutional model that computes
a position sensitive score map shared by every region of interest. \cite{fathi2017semantic}, which is also a fully convolutional approach, learns pixel embedding. Mask R-CNN \cite{he2017mask} extends the FPN model with a branch for predicting masks and introduces new differential cropping operation for both object detection and instance segmentation.

\paragraph{Small objects}
Detecting small objects may be addressed by increasing the input image resolution \cite{chen20153d,liu2016ssd} or by fusing high-resolution features with 
high-dimensional features from the low-resolution image \cite{yang2016exploit,bell2016inside,cao2018feature,DBLP:conf/ivcnz/MenikdiwelaNLS17}. 
This approach of using the higher resolution, however, increases computational overhead and does not address the imbalance between small and large objects.
\cite{li2017perceptual} instead uses a Generative Adversarial Network (GAN) 
to build features in a convolutional network that are indistinguishable between small and large objects in the context of a traffic sign and pedestrian detection. \cite{eggert2017improving} uses different anchor scales based on different resolution layers in a region proposal network. \cite{fang2018small} shifts image features by the correct fraction of the anchor size to cover gaps between them. \cite{chen2016r,ren2018small,cheng2018loco,hu2017finding} add the context when cropping a small object proposal. 

\section{Identifying issues with detecting small objects}
\label{sec:analysis}

In this section, we first overview the MS COCO dataset and the object detection model used in our experiments. We then discuss the issues of the MS COCO dataset and the anchor matching process used in training, that contributes to the difficulty of small object detection. 

\subsection{MS COCO}

We experiment with the MS COCO Detection dataset \cite{lin2014microsoft}. The MS COCO 2017 Detection dataset contains 118,287 images for training, 5,000 images for validation and 40,670 test images. 
860,001 and 36,781 objects from 80 categories are annotated with ground-truth bounding boxes and instance masks.

In the MS COCO detection challenge, the primary evaluation metric is the average precision (AP). In general, AP is defined as the average of ratios of true positives to all positives, for all recall values. 
Because an object needs to be both located and correctly classified, a correct classification is only counted as a true positive detection if the predicted mask or bounding box has an intersection-over-union (IoU) higher than 0.5. 
The AP scores are averaged across the 80 categories and ten IoU thresholds, evenly distributed between 0.5 and 0.95. The metrics also include AP measured across different object scales. In this work, our primary interest is the AP on small objects.

\subsection{Mask R-CNN}

For our experiments, we use the Mask R-CNN implementation from \cite{Detectron2018} with a ResNet-50 backbone 
and adapt the linear scaling rule proposed in \cite{Goyal2017AccurateLM} for setting learning hyperparameters. We use a shorter training schedule than the baselines in \cite{Detectron2018}. We train our models for 36k iterations distributed over four GPUs, using a base learning rate of 0.01. For optimization, we use stochastic gradient descent with the momentum set to 0.9 and weight decay with the coefficient set to 0.0001. The learning rate is scaled down with a factor of 0.1 twice during training, after 24k and 32k iterations. 
All the other parameters are kept as in the baseline Mask R-CNN+FPN+ResNet-50 configuration from \cite{Detectron2018}.

The region proposal stage of the network is particularly important in our investigation. We are using a feature pyramid network (FPN) for generating object proposals \cite{lin2017feature}. It predicts object proposals relative to fifteen anchor boxes from five scales (${32^{2}, 64^{2}, 128^{2}, 256^{2}, 512^{2}}$) and three aspect ratios (${1, 0.5, 2}$). 
An anchor receives a positive label if it has an IoU higher than 0.7 against any ground-truth box, or if it has the highest IoU against a ground-truth bounding box.

\begin{table}[t]
\centering
\begin{tabular}{c || c | c | c | c | c | c}
       & Object  & Images  & Total   & Matched & Average  & Average \\
       & Count   &         & Object  & Anchors & matching & max \\
       &         &         & Area    &         & anchors  & IoU \\ \toprule
small  & 41.43\% & 51.82\% & 1.23\%  & 29.96\% & 1.00     & 0.29 \\ 
medium & 34.32\% & 70.07\% & 10.18\% & 25.54\% & 1.03     & 0.57 \\ 
large  & 24.24\% & 82.28\% & 88.59\% & 44.49\% & 2.54     & 0.66 \\ 
\end{tabular}
\caption{The MS COCO dataset objects statistics with respect to matched anchors in Mask-RCNN based on RPN.}
\label{table:mscocostats}
\end{table}

\begin{figure}[ht!]
    \centering
    \begin{minipage}{0.9\textwidth}
    \begin{minipage}{\textwidth}
    \includegraphics[width=0.49\textwidth]{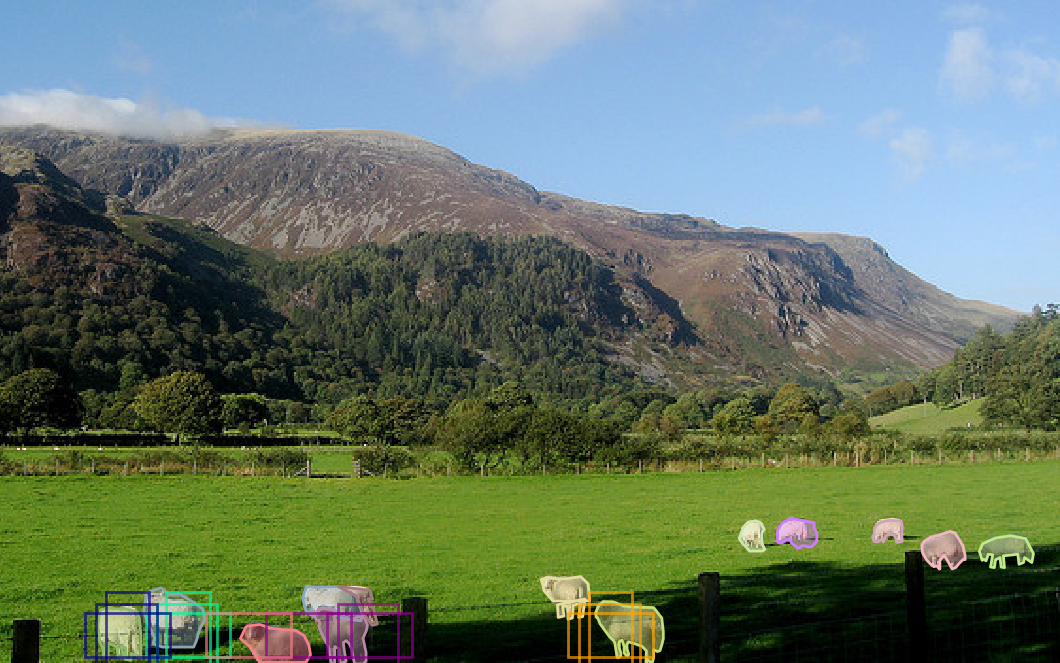}
    \includegraphics[width=0.49\textwidth]{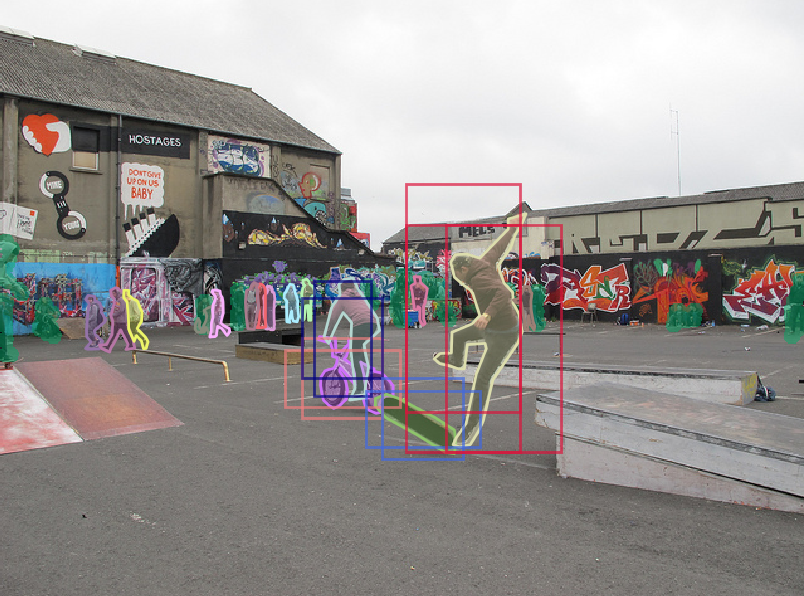}
    \end{minipage}

    \begin{minipage}{\textwidth}
    \includegraphics[width=0.49\textwidth]{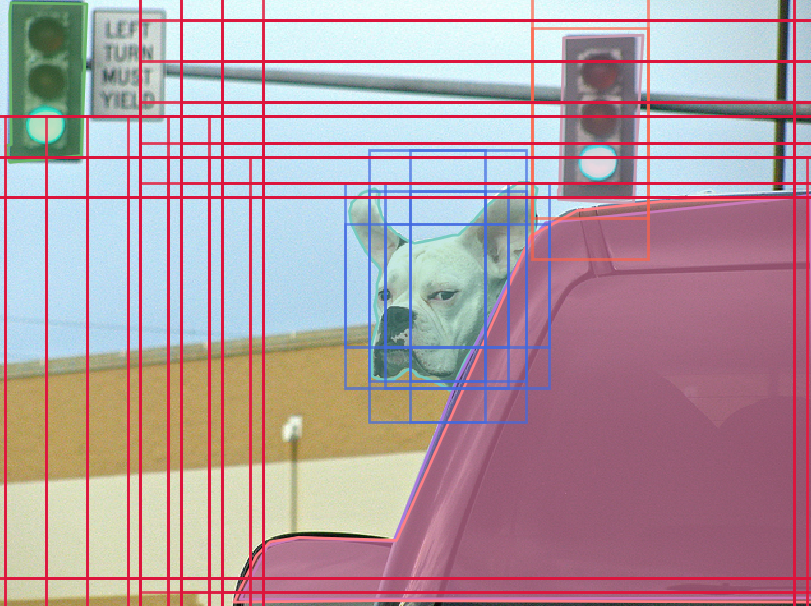}
    \includegraphics[width=0.49\textwidth]{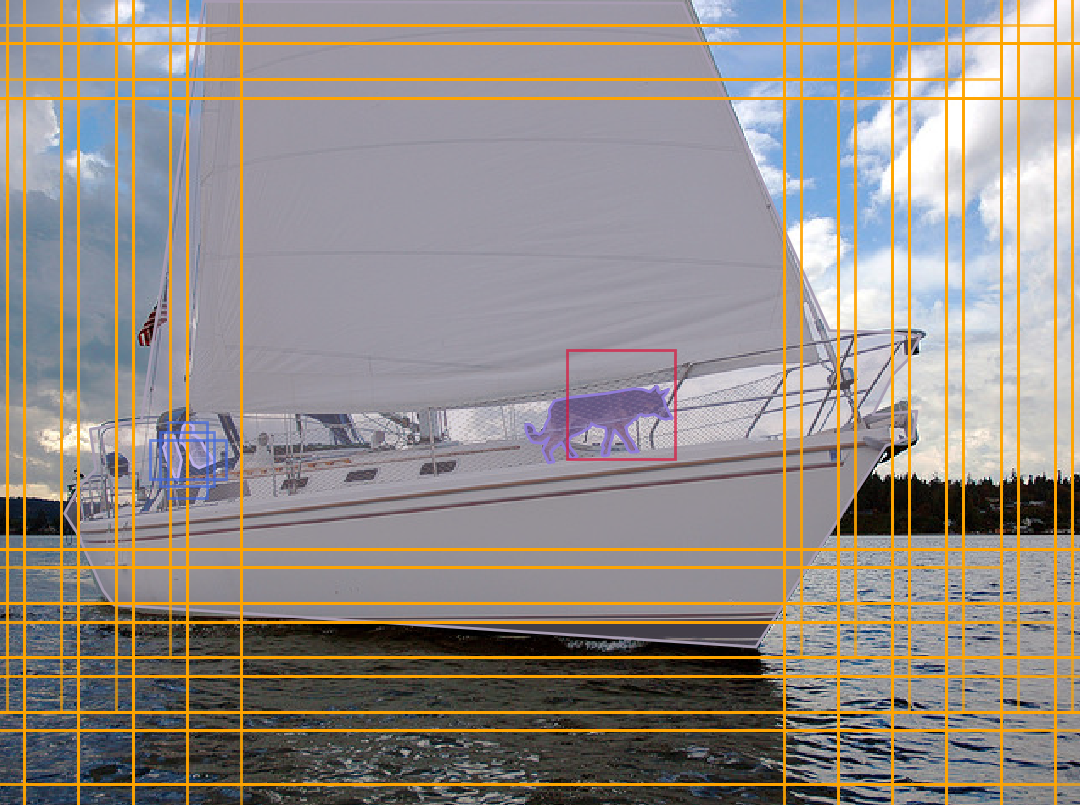}
    \end{minipage}

    \begin{minipage}{0.3\textwidth}
    \includegraphics[width=\columnwidth]{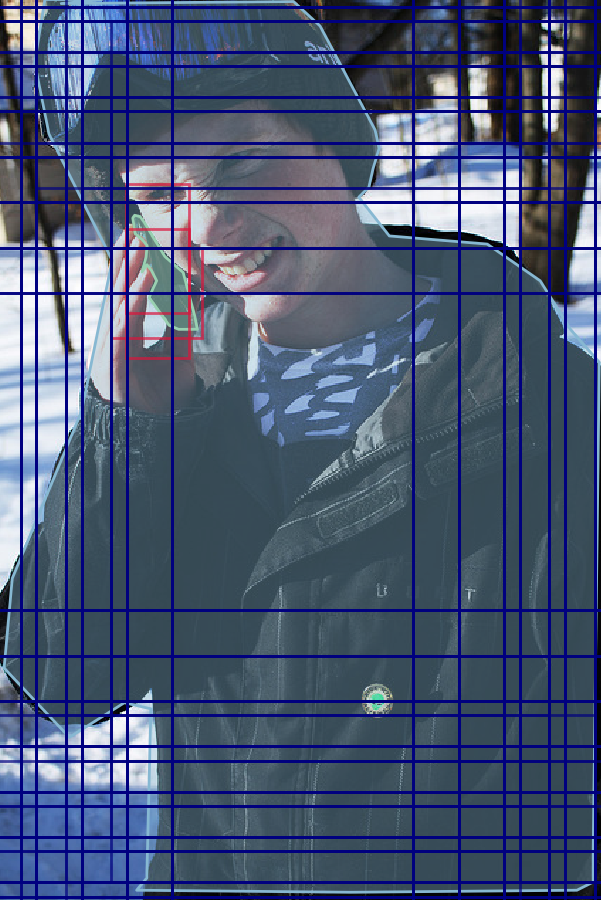}
    \end{minipage}
    \begin{minipage}{0.66\textwidth}
    \caption{Anchors positively assigned (IoU $>$ 0.5) to ground truth objects during training in the Mask-RCNN network. It is necessary to enforce at least one anchor to be positively assigned to every ground truth object even if IoU is below threshold. Otherwise, as on the top 2 images, most of the small objects would be missed and never trained. The other images are verifying the problem we are addressing. The large objects have many much more positively assigned anchors than smaller objects.}
    \label{fig:realAnchors}
    \end{minipage}
    \end{minipage}
\end{figure}

\subsection{Small object detection by Mask R-CNN on MS COCO}

In MS COCO, 41.43\% of all the objects appearing in the training set are small, while only 34.4\% and 24.2\% are medium and large objects respectively. On the other hand, only about half of the training images contain any small objects, while 70.07\% and 82.28\% of training images contain medium and large objects respectively. see \textit{Object Count} and \textit{Images} in the Table~\ref{table:mscocostats}. This confirms the first issue behind the problem of small object detection: there are just fewer examples with small objects.

The second issue is immediately apparent by considering the \textit{Total Object Area} for each size category.
A mere $1.23\%$ of the annotated pixels belong to small objects. Medium sized objects take up already more than eight times more area, $10.18\%$ of the total annotated pixels, while the majority of pixels, $82.28\%$ are labeled as parts of the large objects. Any detector trained on this dataset does not see enough cases of small objects, both across images and across pixels.

As described earlier in this section, each predicted anchor from the region proposal network receives
a positive label if it has the highest IoU with a ground-truth bounding box
or if it has an IoU higher than 0.7 for any ground truth box. 
This procedure highly favors large objects, as 
a large object spanning multiple sliding-window locations often has a high IoU with many anchor boxes, while 
a small object may only be matched with a single anchor box with a low IoU. As listed in Table~\ref{table:mscocostats}, only 29.96\% of positively matched anchors are paired with small objects, while 44.49\% of positively matched anchors with large objects. From the other perspective, it implies that 
there are 2.54 matched anchors per large object, while only one matched anchor per small object. 
Furthermore, 
as the \textit{Average Max IoU} metric reveals, even the best matching anchor box of a small object has a low IoU value typically. The average max IoU for small objects is only 0.29, while medium and large objects have their best matching anchors at around two times higher IoU, 0.57 and 0.66, respectively. We illustrate this phenomenon in fig.~\ref{fig:realAnchors} by visualizing a few examples. These observations suggest that small objects contribute much less 
to computing 
the region proposal loss, which biases the entire network to favor large and medium objects.


\section{Oversampling and Augmentation}
\label{sec:method}

We are improving the performance of object detectors on small objects by explicitly addressing the small object related issues of the MS COCO dataset that we outlined in the previous section. In particular, we over-sample images containing small objects and perform small object augmentation to encourage a model to focus more on small objects. Although we evaluate the proposed approach using Mask R-CNN
it is generally usable with any other 
object detection network or framework, as 
both oversampling and augmentation are done 
as data preprocessing.

\paragraph{Oversampling}

We address the issue of relatively fewer images containing small objects by oversampling those images during training~\cite{buda2017systematic}. It is an effortless and straight-forward way to alleviate this problem of the MS COCO dataset and improve performance on small object detection. In the experiments, we vary the oversampling rate and investigate the effect of oversampling not only on small object detection but also on detecting medium and large objects.

\paragraph{Augmentation}

On top of oversampling we also introduce dataset augmentation focused on small objects. Instance segmentation masks provided in the MS COCO dataset allow us to make a copy of any object from its original location. The copy is then  
pasted to 
different positions. By increasing the number of small objects in each image, the number of matched anchors increases.
This, in turn, improves the contribution of small objects to computing the loss function of the RPN during training.

Before pasting the object to a new location, we apply random transformations on it. We scale the objects by changing the object size $\pm 20\%$ and rotate it $\pm 15^{\circ}$.
We only consider non-occluded objects, as pasting disjoint segmentation masks with unseen parts in-between often results in less realistic images.
We ensure that the newly pasted object does not overlap with any existing object and is at least five pixels away from the image boundaries.

In Fig.~\ref{fig:schematic_anchors}, we graphically illustrate the proposed augmentation strategy and how it increases the number of matched anchors during training, leading to a better detector of small objects.


\section{Experimental Setup}

\subsection{Oversampling}

In the first set of experiments, we investigate the effect of 
oversampling images containing small objects.
We vary the oversampling ratio between two, three and four. Instead of actual stochastic oversampling, we create multiple copies of images with small objects offline for efficiency.

\subsection{Augmentation}

In the second set of experiments, we investigate the effects of using augmentation on small object detection and segmentation. We copy and paste all small objects in each image once. We also oversample images with small objects to study the interaction between the oversampling and augmentation strategies.

We test three settings. In the first setting, we replace each image with small objects by the one with copy-pasted small objects. In the second setting, we duplicate these augmented images to mimic oversampling. In the final setup, we keep both the original images and augmented images, which is equivalent to oversampling the images with small objects by the factor of two, while augmenting the duplicated copies with more small objects.

\subsection{Copy-Pasting Strategies}

There are different ways to copy-pasting small objects. We consider three separate strategies. First, we pick one small object in an image and copy-paste it multiple times in random locations. Second, we choose numerous small objects and copy-paste each of these exactly once in an arbitrary position. Lastly, we copy-paste all small objects in each image multiple times in random places. In all the cases, we use the third setting of augmentation above; that is, we keep both the original image and its augmented copy.

\subsection{Pasting Algorithms}

When pasting a copy of a small object, there are two things to consider. First, we must decide whether a pasted object would overlap with any other object. Although we choose not to introduce any overlap, we experimentally verify whether it is a good strategy. Second, it is a design choice whether to perform an additional procedure to smooth the edge of a pasted object. We experiment whether Gaussian blurring of the boundary with varying filter sizes could help compared to no further processing. 


\section{Result and Analysis}

\begin{table}[t]
\centering
\begin{tabular}{c||c|c|c|c||c|c|c|c}
                                      & \multicolumn{4}{c||}{Segmentation AP} & \multicolumn{4}{c}{Detection AP} \\ 
                                      & small   & medium   & large  & all    & small  & medium  & large  & all   \\ 
                                      \toprule
\multicolumn{1}{c||}{baseline}        & 0.113   & 0.300    & 0.418  & 0.28   & 0.167  & 0.329   & 0.393  & 0.303 \\ \midrule
\multicolumn{1}{c||}{oversampling 2$\times$} & 0.120   & 0.299    & 0.409  & 0.279  & 0.173  & 0.328   & 0.387  & 0.304 \\ \midrule
\multicolumn{1}{c||}{oversampling 3$\times$} & \textbf{0.123}   & 0.300    & 0.404  & {0.279}  & \textbf{0.177}  & 0.329   & 0.382  & \textbf{0.305} \\ \midrule
\multicolumn{1}{c||}{oversampling 4$\times$} & 0.120   & 0.299    & 0.398  & 0.276  & 0.174  & 0.329   & 0.374  & 0.302 \\ 
\end{tabular}
\caption{Experiments with different oversampling ratios. We observe that oversampling helps regardless of the ratio for detecting small objects. The ratio allows us to make a trade-off between small and large objects.}
\label{table:results_oversampling}
\end{table}

\subsection{Oversampling}

By sampling the small object images more often during training (see Table~\ref{table:results_oversampling}), AP on both small object segmentation and detection can be improved. The most gain is observed with 3$\times$ oversampling, which increases AP for small objects by $1\%$ (corresponding to a relative improvement of $8.85\%$). While performance on the medium object scale is less affected, large object detection and segmentation performance consistently suffer from oversampling, implying that the ratio must be chosen based on the relative importance between small and large objects.

\begin{table}[t]
\centering
\begin{tabular}{c||c|c|c|c||c|c|c|c}
                                        & \multicolumn{4}{c|}{Segmentation AP}            & \multicolumn{4}{c|}{Detection AP}                \\ 
                                        & small          & medium & large & all           & small          & medium & large & all            \\ 
                                        \toprule
\multicolumn{1}{c||}{baseline}          & 0.113          & 0.300  & 0.418 & 0.28          & 0.167          & 0.329  & 0.393 & 0.303          \\ \midrule
\multicolumn{1}{c||}{aug}               & 0.108          & 0.299  & 0.422 & 0.278         & 0.161          & 0.328  & 0.4   & 0.302          \\ \midrule
\multicolumn{1}{c||}{aug+oversample 2x} & 0.117          & 0.300  & 0.406 & 0.277         & 0.168          & 0.333  & 0.387 & 0.302          \\ \midrule
\multicolumn{1}{c||}{original+aug}      & \textbf{0.124} & 0.301  & 0.41  & \textbf{0.28} & \textbf{0.179} & 0.329  & 0.386 & \textbf{0.304} \\ 
\end{tabular}
\caption{Augmentation experiments. The best performance, in terms of both small objects and overall, is achieved when the original images with small objects and their copy with copy-pasted small objects are used for training. }
\label{table:results2}
\end{table}

\subsection{Augmentation}

In Table~\ref{table:results2}, we present the results using different combinations of the proposed augmentation and oversampling strategy. When we replace each image with small objects by its copy that contains more small objects (the second row), the performance degraded notably. When we oversampled these augmented images by the factor of two, the segmentation and detection performance on the small objects regained its loss, although the overall performance was still worse than the baseline. When we evaluated this model on
an augmented validation set, instead of the original one, we, however, saw a 38\% increase in the small object augmentation performance (0.161),  suggesting that the trained model effectively overfit to ``pasted'' small objects but not necessarily to the original small objects. 
We believe this is due to 
the artifacts from pasting, such as 
imperfect object masks and brightness differences from the background,
that are relatively easy for a neural network to spot. 
The best results were achieved by combining oversampling and doing augmentation with a probability of $p=0.5$ (\textit{original+aug}) with the ratio of original to augmented small objects is 2:1. This setting yielded better results than oversampling alone, confirming the effectiveness of the proposed strategy of pasting small objects.

\subsection{Copy-Pasting strategies}

\begin{table}[t]
\centering
\begin{tabular}{l||c|c|c|c||c|c|c|c}
                               & \multicolumn{4}{c||}{Segmentation AP}    & \multicolumn{4}{c}{Detection} \\ 
                               & small          & medium & large & all   & small & medium & large & all   \\ 
                               \toprule
\multicolumn{1}{l||}{baseline} & 0.113          & 0.300  & 0.418 & 0.280 & 0.167 & 0.329  & 0.393 & 0.303 \\ \midrule
\multicolumn{1}{l||}{1$\times$ pasted} & \textbf{0.122} & 0.303  & 0.405 & \textbf{0.281} & \textbf{0.174} & 0.333  & 0.384 & \textbf{0.307} \\ \midrule
\multicolumn{1}{l||}{2$\times$ pasted} & 0.122          & 0.300  & 0.408 & 0.279 & 0.175 & 0.330  & 0.382 & 0.304 \\ \midrule
\multicolumn{1}{l||}{3$\times$ pasted} & 0.121          & 0.301  & 0.403 & 0.280 & 0.175 & 0.330  & 0.386 & 0.305 \\ \midrule
\multicolumn{1}{l||}{4$\times$ pasted} & 0.120          & 0.298  & 0.403 & 0.277 & 0.174 & 0.328  & 0.383 & 0.304 \\ \midrule
\multicolumn{1}{l||}{5$\times$ pasted} & 0.116          & 0.298  & 0.405 & 0.277 & 0.171 & 0.328  & 0.386 & 0.304 \\ 
\end{tabular}
\caption{Copy-pasting of a single object. We observe that it is often best to copy-paste a single object only a few times (1$\times$ or 2$\times$,) especially to achieve the high overall performance.}
\label{table:results_random_single}
\end{table}

\paragraph{Copy-pasting of a single object}

In Table~\ref{table:results_random_single}, we see that copy-pasting a single object results in a better model on small objects, however, at the cost of a small performance drop on large images. These results are also better than two times oversampling in itself. The performance, however, peaks already at one or two pastes. Adding the same object more times does not yield any performance improvement.

\begin{table}[t]
\centering
\begin{tabular}{l||c|c|c|c||c|c|c|c}
                               & \multicolumn{4}{c||}{Segmentation AP} & \multicolumn{4}{c}{Detection} \\ 
                               & small   & medium   & large  & all    & small & medium & large & all   \\ 
                               \toprule
\multicolumn{1}{l||}{baseline} & 0.113   & 0.300    & 0.418  & 0.280  & 0.167 & 0.329  & 0.393 & 0.303 \\ \midrule
\multicolumn{1}{l||}{1$\times$ pasted} & 0.122   & 0.303    & 0.405  & 0.281  & 0.174 & 0.333  & 0.384 & 0.307 \\ \midrule
\multicolumn{1}{l||}{2$\times$ pasted} & 0.121   & 0.300    & 0.406  & 0.279  & 0.175 & 0.329  & 0.385 & 0.305 \\ \midrule
\multicolumn{1}{l||}{3$\times$ pasted} & \textbf{0.124}   & 0.303    & 0.411  & \textbf{0.281}  & \textbf{0.178} & 0.331  & 0.387 & \textbf{0.306} \\ \midrule
\multicolumn{1}{l||}{4$\times$ pasted} & 0.120   & 0.299    & 0.403  & 0.279  & 0.179 & 0.329  & 0.382 & 0.305 \\ \midrule
\multicolumn{1}{l||}{5$\times$ pasted} & 0.119   & 0.301    & 0.412  & 0.279  & 0.176 & 0.330  & 0.389 & 0.305 \\ 
\end{tabular}
\caption{Copy-pasting of multiple objects. Compared to copy-pasting of a single object, it is better to make more copies (3$\times$).}
\label{table:results_random_multiple}
\end{table}

\paragraph{Copy-pasting of multiple objects}

As it can be seen in Table~\ref{table:results_random_multiple}, it is better to copy-paste multiple small objects per image than to copy-paste only a single object. 
In this case, we see the benefits of pasting up to three times per object.

\begin{table}[t]
\centering
\begin{tabular}{l||c|c|c|c||c|c|c|c}
                               & \multicolumn{4}{c||}{Segmentation AP} & \multicolumn{4}{c}{Detection} \\ 
                               & small   & medium   & large  & all    & small & medium & large & all   \\ 
                               \toprule
\multicolumn{1}{l||}{baseline} & 0.113   & 0.300    & 0.418  & 0.280  & 0.167 & 0.329  & 0.393 & 0.303 \\ \midrule
\multicolumn{1}{l||}{1$\times$ pasted}      & \textbf{0.124}   & 0.301    & 0.410  & \textbf{0.280}  & \textbf{0.179} & 0.329  & 0.386 & \textbf{0.304} \\ \midrule
\multicolumn{1}{l||}{2$\times$ pasted}  & 0.119   & 0.299    & 0.410  & 0.279  & 0.179 & 0.328  & 0.388 & 0.304 \\ \midrule
\multicolumn{1}{l||}{3$\times$ pasted}  & 0.113   & 0.299    & 0.405  & 0.276  & 0.167 & 0.330  & 0.383 & 0.302 \\ 
\end{tabular}
\caption{Copy-pasting of all small objects. It is best to make only one copy of every small object, and this strategy does not outperform the strategy of copy-pasting multiple (but not all) objects many times.}
\label{table:results_random_all}
\end{table}

\paragraph{Copy-pasting of all small objects}

Finally, Table \ref{table:results_random_all} lists the results where all the small objects in each image are copy-pasted. We found the best results concerning both the segmentation and detection at augmenting with all the objects once. We suspect two possible causes behind this. First, By having multiple copies of all small objects the ratio of original to pasted small objects rapidly decreases. Second, the number of objects in each image multiplies, and this causes a more considerable mismatch between training and test images.

\subsection{Pasting Algorithms}

As shown in the Table~\ref{table:results_params}, pasting randomly into images without considering what other objects already occupy areas leads to inferior performance on small images. It justifies our design choice to avoid any overlap between a pasted object and existing objects. Further, Gaussian blurring of the edge of a pasted object did not show any improvement, suggesting that it is better to paste an object as it is, unless with a more sophisticated strategy of fusing in the object.

\begin{table}[t]
\centering
\begin{tabular}{l||c|c|c|c||c|c|c|c}
                                   & \multicolumn{4}{c||}{Segmentation AP}    & \multicolumn{4}{c}{Detection} \\ 
                                   & small          & medium & large & all   & small & medium & large & all   \\ 
                                   \toprule
\multicolumn{1}{l||}{original+aug} & \textbf{0.124} & 0.301  & 0.41  & \textbf{0.28}  & \textbf{0.179} & 0.329  & 0.386 & \textbf{0.304} \\ \midrule
\multicolumn{1}{l||}{overlapping}  & 0.118          & 0.301  & 0.411 & 0.279 & 0.173 & 0.331  & 0.389 & 0.305 \\ \midrule
\multicolumn{1}{l||}{blur 3}       & 0.117          & 0.299  & 0.408 & 0.278 & 0.17  & 0.329  & 0.389 & 0.304 \\ \midrule
\multicolumn{1}{l||}{blur 5}       & 0.119          & 0.3    & 0.408 & 0.279 & 0.173 & 0.328  & 0.383 & 0.303 \\ 
\end{tabular}
\caption{Results with different pasting algorithms. We observe that it is crucial not to introduce any overlap when copy-pasting a small object, and that it is not advisable to Gaussian-blur the edge of a pasted object.}
\label{table:results_params}
\end{table}

\section{Conclusion}

We investigated the problem of small object detection. We showed that one of the factors behind the poor average precision for small objects is the lack of representation of small objects in a training data. This is especially true with the existing state-of-the-art object detector which requires the presence of enough objects for predicted anchors to match during training. 
We proposed two strategies 
for augmenting the original MS COCO database to overcome the issue. First, we show the performance on small objects can easily improve by oversampling images containing small objects during training.
Second, we propose an augmentation algorithm based on copy-pasting small objects. 
Our experiments proved a 9.7\% relative improvement for the instance segmentation and 7.1\% for object detection for small objects compared to the current state of the art, obtained by Mask R-CNN, on MS COCO. The proposed set of augmentation methods offers the trade-off between the quality of predictions for small and large objects, as verified by the experiments. 

\bibliographystyle{splncs04}
\bibliography{egbib}

\end{document}